\documentclass[conference]{IEEEtran}
\usepackage{cite}
\usepackage{amsmath,amssymb,amsfonts}
\usepackage{graphicx}
\usepackage{booktabs}
\usepackage{xcolor}
\usepackage{multirow}
\usepackage{microtype}

\usepackage{svg}

\newcommand{\figpdf}[3]{
\IfFileExists{#1}{\includegraphics[width=#2]{#1}}{%
\fbox{\parbox[c][#3][c]{\dimexpr#2-2\fboxsep-2\fboxrule}%
{\centering\footnotesize\textcolor{gray}{[upload \texttt{\detokenize{#1}}]}}}}}

\begin{document}

\title{\huge{RingNav: Task-Shaped Circular Memory for Visual Heading}}
\author{Naitri Rajyaguru$^{*}$, Aryan Tajne$^{*}$, Stephen Jardim, Cornelia Fermuller, Yiannis Aloimonos \\
\footnotesize{$^{*}$Equal Contrbution}\\[6pt]
\normalsize{University of Maryland, College Park}}
\maketitle

\begin{abstract}
Tiny robots have limited computation and memory, yet visual navigation
systems often maintain more state than a task requires. We ask whether
persistent memory can instead be shaped by the task. We present
\textbf{RingNav}, a \(10.6\)k--parameter visual heading estimator that
constrains its memory to the circular manifold \(S^1\). RingNav learns only
the turn between consecutive images, while heading is composed directly on
the circle. On the TartanAir Visual SLAM Challenge,
RingNav reduces mean absolute turn error by \(27\%\) and
accumulated heading error over $4$ sec by \(15\%\) relative to a
similarly sized learned-memory model.
The structured update also gives exactly zero turn for identical frames and
exact cancellation under sequence reversal. Without fine-tuning, RingNav
transfers to real-world drone experiments and continues estimating heading under severe visual degradation. 
RingNav runs at $27.3$Hz on Raspberry Pi Zero 2W and can be deployed on an ultra-low power GAP8 processor at $33.4$Hz.
\end{abstract}

\section{Introduction}

Autonomous visual navigation remains difficult to deploy on small robots because computation and memory are limited. Visual odometry (VO) estimates camera motion from an image sequence, while simultaneous localization and mapping (SLAM) additionally maintains a representation of the environment for localization \cite{engel2017dso,campos2021orbslam3,wang2021tartanvo, teed2023deep,teed2021droid}. These systems recover rich motion and scene information, even when a task may require only a small part of that state.

Recent work has reduced the cost of visual navigation for resource-constrained robots through smaller estimators and task-specific representations \cite{palossi2019pulpdronet,marchei2026tinydevo, vandijk2024route,ou2026efficient,gattaux2026antflie, hagenaars2026alleyes}. These methods reduce the cost of estimating or storing navigation information. A second source of complexity is what the robot must remember over time. Different behaviors require different information, ranging from no persistent state for reactive control to full pose for metric navigation. This idea is consistent with purposive and task-driven estimation, which seek representations sufficient for the behavior rather than recovering the full physical state \cite{sanket2018gapflyt,pacelli2019task,sakcak2024minimal}.

We call this principle \emph{task-shaped memory}: when the state to be remembered and its update rule are known from the task, learning is needed only to estimate how that state changed from the observations. Different tasks can therefore require different memory structures, as shown in Table~\ref{tab:task_memory}.

\begin{figure}[t!]
\centering
 \centering
    \includegraphics[width=\columnwidth]{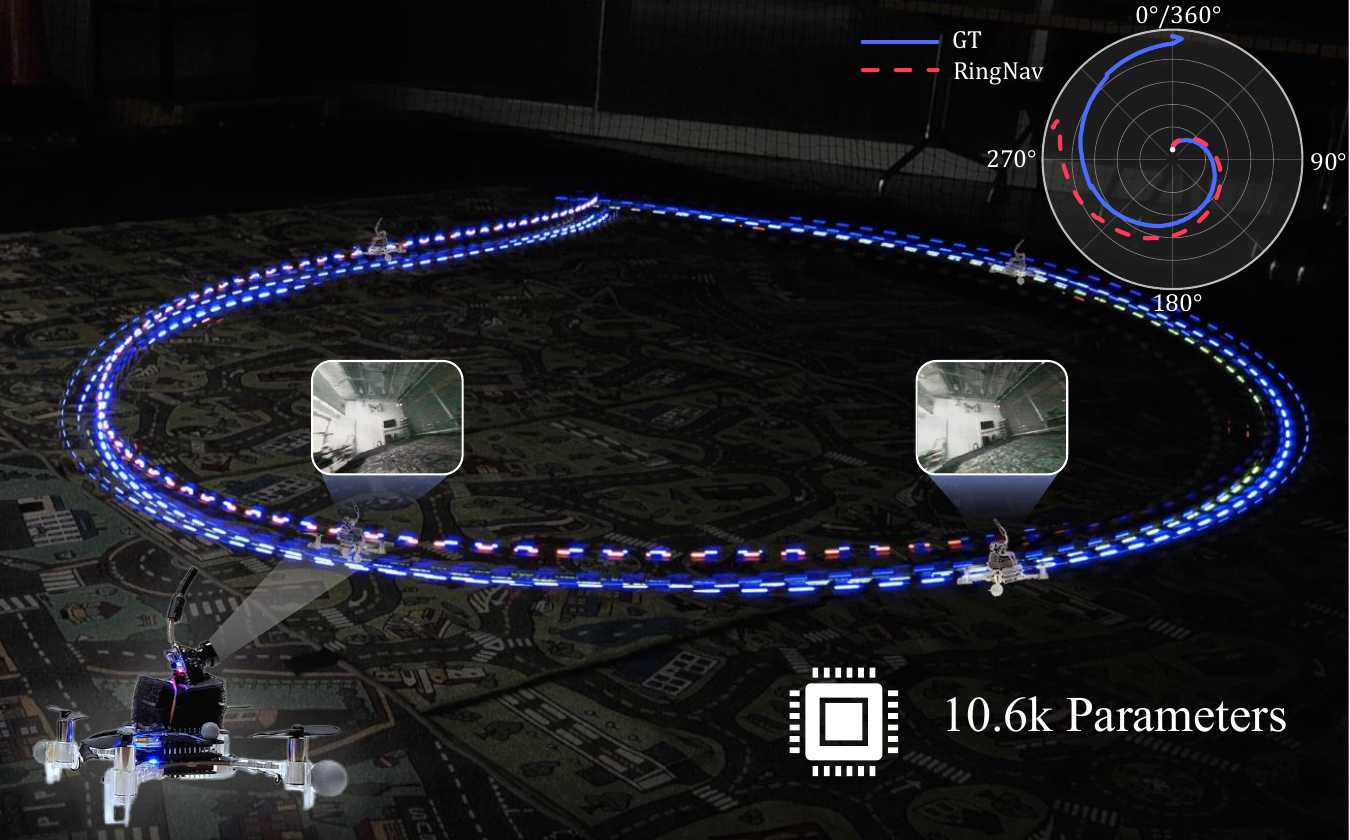}
\caption{\textbf{Task-shaped memory for visual heading.}
RingNav estimates the visual turn between consecutive images and maintains
heading as a persistent circular state. On a real Crazyflie flight, its
heading estimate follows motion-capture ground truth around the trajectory.
The estimator uses 10.6k trainable parameters.}
\label{fig:hero}
\end{figure}

\begin{table}[t!] \caption{\textbf{Task-selected persistent state.} Different behaviors require different information to persist over time.}
\label{tab:task_memory}
\centering \scriptsize \renewcommand{\arraystretch}{1.10} \setlength{\tabcolsep}{2.7pt} \begin{tabular*}{\columnwidth}{@{\extracolsep{\fill}}llll@{}} \toprule \textbf{Task dependence} & \textbf{State} & \textbf{DoF} & \textbf{State update}\\ \midrule Heading & \(S^1\) & 1 & circular rotation\\ Heading + progress & \(S^1\times\mathbb{R}\) & 2 & rotation + addition\\ Heading + cyclic progress & \(T^2=S^1\times S^1\) & 2 & two circular rotations\\ 3D bearing & \(S^2\) & 2 & \(SO(3)\) action\\ Planar pose & \(SE(2)\) & 3 & rigid-body composition\\ 3D pose & \(SE(3)\) & 6 & rigid-body composition\\ \bottomrule \end{tabular*} \end{table} These examples illustrate the design principle; this work experimentally studies only heading on \(S^1\).

For heading, the state has one degree of freedom but is periodic: after a full rotation, the robot faces the same direction again. Its possible values therefore form the circle \(S^1\). A recurrent network can instead encode heading in a latent hidden state \(h_t\in\mathbb{R}^n\), whose representation and update are learned from data. Task-shaped memory instead fixes the heading state to \(S^1\) and its update to circular rotation, leaving only the visual turn between consecutive images to be learned. 

This leads to \textbf{RingNav}, a \(10.6\)k-parameter monocular heading estimator. A lightweight image-pair network estimates the visual turn \(\Delta\psi_t\) between consecutive frames, and this turn directly rotates the stored heading on \(S^1\). The estimator is further constrained so that identical images produce exactly zero turn and reversing the image order reverses the predicted turn. To isolate the effect of these choices, we compare RingNav with similarly sized alternatives that learn the persistent state, its update, or both.

\subsection{Key Contributions} \begin{itemize} \item We present \textbf{RingNav}, a 10.6k-parameter monocular heading estimator that learns the turn between images while storing heading on \(S^1\) and updating it directly with the predicted turn, rather than learning the persistent state and its update.
 \item We constrain the visual turn estimator so that identical frames produce exactly zero predicted turn and reversing the image order exactly reverses the prediction. \item We provide controlled evaluation under unseen, degraded, repeated, and missing visual information, together with comparisons to rotation-only and full visual-motion estimators, zero-shot transfer to real Crazyflie flight data, and edge-runtime evaluation. \end{itemize}
\section{Related Work}

\textbf{Visual Rotation and Heading Estimation.}
Heading is commonly propagated from inertial measurements by integrating
gyroscope angular rate, while magnetometers can provide a yaw reference.
Gyroscope integration accumulates drift over time, and magnetometer-based
heading can be degraded by local magnetic disturbance
\cite{li2024robust,cui2024vehicle}. Vision provides a scene-derived cue that
is complementary to inertial heading propagation. Vision-only attitude and
angular-rate estimation has also been demonstrated for flight control
\cite{hagenaars2026alleyes}.

Visual compasses estimate heading from appearance
\cite{labrosse2006visualcompass,sturm2009visualcompass}, while rotation-only
visual estimators recover camera rotation without estimating translation or
an explicit 3D map
\cite{mariottini2012visualcompass,hartmann2015visualgyro}. Feature-based
visual rotation can be accurate, but requires feature extraction, matching,
and geometric estimation \cite{hartmann2015visualgyro}.
Rotation has also been estimated from event cameras
\cite{gallego2017angular} and by combining learned features with direct
alignment \cite{berenguel2023visualgyro}. Full visual odometry estimates both
rotation and translation
\cite{wang2021tartanvo,teed2023deep,teed2021droid}, and methods such as DPVO
and DROID-SLAM combine learned components with geometric optimization
\cite{teed2023deep,teed2021droid}. Task-specific sensing has also been used to improve visual odometry
\cite{shah2024codedvo,rajyaguru2026polardepth}.

Prior work has also used forward and backward motion estimates
\cite{sardana2019joint,sardana2023improving}, while recent relative-pose work
has enforced inverse consistency when the order of two input images is
reversed \cite{liu2025equipose}.  RingNav imposes the corresponding property on visual
turn estimation: reversing the image order exactly reverses the predicted
turn. The predicted turn is then used to update the persistent \(S^1\)
heading state.

\textbf{Visual Navigation on Resource-Constrained Robots.}
Small robots have motivated compact learned navigation
\cite{palossi2019pulpdronet}, event-based visual odometry on ultra-low-power
hardware \cite{marchei2026tinydevo}, compact route memories
\cite{vandijk2024route}, visual homing and path integration
\cite{ou2026efficient}, and teach-and-repeat navigation
\cite{gattaux2026antflie}. These works reduce the computation needed for
navigation or reduce the information that must be estimated. More broadly,
task-driven estimation seeks representations sufficient for a behavior rather
than recovering the full physical state
\cite{pacelli2019task,sakcak2024minimal}. RingNav asks a related question:
if heading already has a known circular structure, does the robot need to
learn how to store and update it?

\textbf{Structured Representations of Heading.}
Biological heading systems provide an example of memory with a circular
structure. In insects, heading-related neural activity forms a persistent
ring-like representation that moves as the animal turns
\cite{seelig2015neural,green2017angular,kim2017ring,
turnerevans2017angular}, and circular organization has also been observed in
mammalian head-direction populations
\cite{chaudhuri2019intrinsic}. Circular task states have also been built into
recurrent networks \cite{pollock2020engineering}, and ring-attractor models
have been used for robotic orientation and navigation
\cite{robinson2022orientation,riveroortega2023ring}.

Structured representations and composition rules for orientation and pose are
also widely used in robotic state estimation \cite{sola2018micro}. RingNav
uses the known circular structure as the persistent heading state and fixes
its rotation update, leaving only the visual turn between images to be
learned.

\section{Method}
\label{sec:method}

Let \(I_t\) denote the \(t\)-th sampled image, and let \(\psi_t\) denote its
heading about the vertical axis. The turn between successive observations is

\vspace{-4pt}
\begin{equation}
\Delta\psi_t =
\operatorname{wrap}\!\left(\psi_{t+1}-\psi_t\right),
\label{eq:gt_turn}
\end{equation}

where \(\operatorname{wrap}:\mathbb{R}\rightarrow[-\pi,\pi)\).

\begin{figure*}[t]
    \centering
    \includegraphics[
        width=0.9\textwidth,
        trim={0cm 0.25cm 0cm 0.25cm},
        clip
    ]{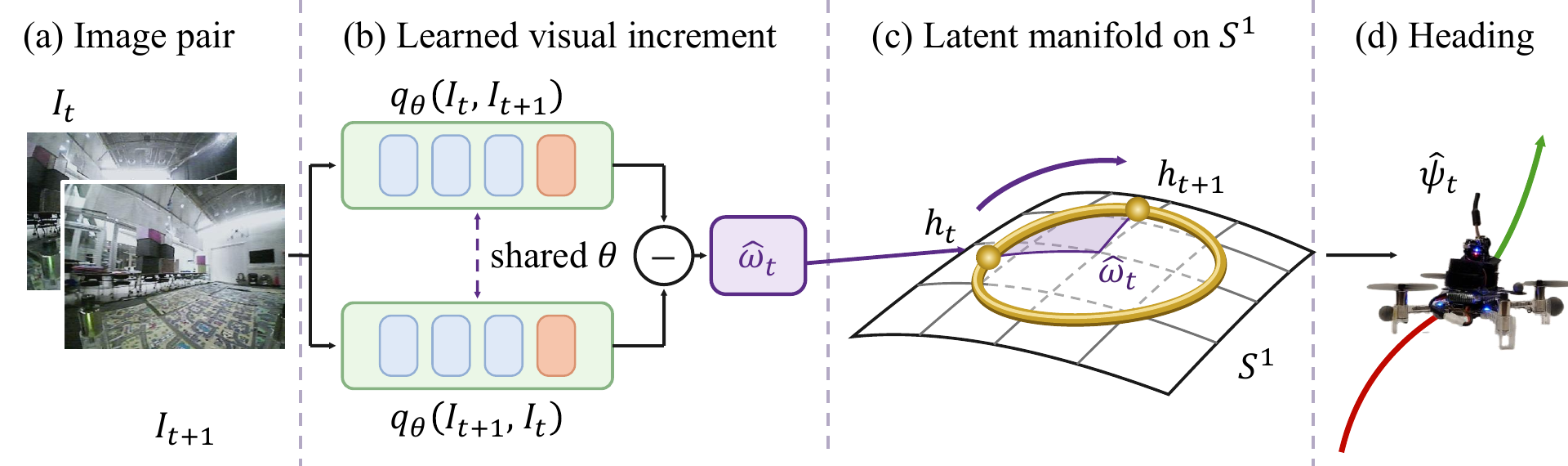}
    \caption{\textbf{RingNav pipeline.}
    A shared image-pair network processes consecutive observations in both
    temporal orders. Their half-difference gives the visual increment
    \(\widehat{\Delta\psi}_t\), which acts on the latent manifold \(S^1\)
    through \(R(\widehat{\Delta\psi}_t)\), advancing the persistent state
    from \(s_t\) to \(s_{t+1}\) and yielding the updated heading
    \(\hat\psi_{t+1}\).}
    \label{fig:state}
\end{figure*}

RingNav consists of an ordered image-pair mapping \(q_\theta\), an
antisymmetric projection that gives the visual increment
\(\widehat{\Delta\psi}_t\), and a fixed rotation acting on a persistent latent
state \(s_t\in S^1\), as shown in Fig.~\ref{fig:state}.

\subsection{Visual Increment Network}
\label{sec:pair}

For an ordered image pair \((I_a,I_b)\), RingNav defines the scalar mapping
\[
q_\theta:\mathcal I\times\mathcal I\rightarrow\mathbb{R},
\]
where \(\mathcal I\) denotes the image space. Each image is
\(320\times320\) RGB. For the forward ordering \((I_t,I_{t+1})\), the
ImageNet-normalized images are concatenated channel-wise and encoded as

\vspace{-4pt}
\begin{equation}
z_t =
E_\theta\!\left([I_t,I_{t+1}]\right),
\qquad
z_t\in\mathbb{R}^{24}.
\label{eq:encoder}
\end{equation}

The encoder \(E_\theta\) contains four \(3\times3\) convolutions with stride
2, no bias, and \(8\), \(12\), \(16\), and \(24\) output channels, each
followed by batch normalization and ReLU. Global average pooling after the
final block gives \(z_t\). A two-layer MLP with hidden and output width 32
produces

\vspace{-4pt}
\begin{equation}
f_t =
W_2\,\operatorname{ReLU}(W_1z_t+b_1)+b_2,
\qquad
f_t\in\mathbb{R}^{32}.
\label{eq:features}
\end{equation}

A gated scalar head gives the ordered prediction

\vspace{-4pt}
\begin{equation}
q_\theta(I_t,I_{t+1})
=
w_o^\top
\left[
(Af_t+a)\odot\sigma(Bf_t+b)
\right],
\label{eq:score}
\end{equation}

where \(A,B\in\mathbb{R}^{32\times32}\),
\(a,b,w_o\in\mathbb{R}^{32}\), \(\sigma(\cdot)\) is the sigmoid function,
and \(\odot\) denotes element-wise multiplication.

\subsection{Antisymmetric Projection}
\label{sec:skew}

The ordered mapping \(q_\theta\) is not itself constrained to be
antisymmetric. We evaluate the shared network in both temporal orders,

\begin{equation}
q_{f,t}=q_\theta(I_t,I_{t+1}),
\qquad
q_{r,t}=q_\theta(I_{t+1},I_t),
\label{eq:branches}
\end{equation}

and take their antisymmetric component,

\begin{equation}
\widehat{\Delta\psi}_t
=
\frac{1}{2}
\left(q_{f,t}-q_{r,t}\right).
\label{eq:skew}
\end{equation}

Equivalently, for arbitrary \(I_a,I_b\),

\begin{equation}
\widehat{\Delta\psi}(I_a,I_b)
=
\frac{1}{2}
\left[
q_\theta(I_a,I_b)-q_\theta(I_b,I_a)
\right].
\label{eq:skew_general}
\end{equation}

It follows directly that

\begin{equation}
\widehat{\Delta\psi}(I_b,I_a)
=
-\widehat{\Delta\psi}(I_a,I_b),
\qquad
\widehat{\Delta\psi}(I,I)=0.
\label{eq:identities}
\end{equation}

Thus reversing an image pair exactly reverses the predicted increment, while
identical images produce zero increment. In exact arithmetic, the predicted
increments from a sequence and the same sequence traversed in reverse
therefore sum to zero.

The two evaluations share all parameters. Because the images are concatenated
before \(E_\theta\), the forward and reverse orderings require separate
image-pair evaluations.

\subsection{Latent Manifold Dynamics on \(S^1\)}
\label{sec:state}

The persistent latent state is constrained to the unit circle

\begin{equation}
S^1 =
\left\{
s\in\mathbb{R}^{2}:\|s\|_2=1
\right\}.
\label{eq:manifold}
\end{equation}

An estimated heading \(\hat\psi_t\) is represented by

\begin{equation}
s_t =
\begin{bmatrix}
\cos\hat\psi_t\\
\sin\hat\psi_t
\end{bmatrix}
\in S^1.
\label{eq:state}
\end{equation}

The persistent latent state contains two FP32 values, or 8 bytes. The visual
increment acts on \(S^1\) through the planar rotation

\begin{equation}
R(\delta)=
\begin{bmatrix}
\cos\delta & -\sin\delta\\
\sin\delta & \cos\delta
\end{bmatrix}
\in SO(2).
\label{eq:rotation}
\end{equation}

The state transition is

\begin{equation}
s_{t+1}
=
\Pi_{S^1}
\left(
R(\widehat{\Delta\psi}_t)s_t
\right),
\label{eq:update}
\end{equation}

where
\vspace{-2pt}
\begin{equation}
\Pi_{S^1}(x)
=
\frac{x}{\|x\|_2}.
\label{eq:reprojection}
\end{equation}

In exact arithmetic,
\(R(\delta)S^1=S^1\), so the rotation preserves the manifold. The projection
in Eq.~\eqref{eq:reprojection} only corrects finite-precision changes in
vector norm. Equation~\eqref{eq:update} contains no learned parameters.

The heading represented by the latent state is recovered as

\begin{equation}
\hat\psi_t =
\operatorname{atan2}\!\left(s_{t,2},s_{t,1}\right).
\label{eq:heading_readout}
\end{equation}

For an initial latent state \(s_0\in S^1\), successive visual increments
therefore propagate the heading entirely through the fixed circular update.

\subsection{Training Objective}
\label{sec:losses}

Training uses 24-frame windows whose temporal stride is drawn uniformly from
\(\{1,2\}\) for each window. Within a window, \(t\) indexes the sampled
observations, so \(I_t\) and \(I_{t+1}\) denote consecutive samples in that
window. Validation and all evaluations use stride 1.

For angles \(a\) and \(b\), define the circular error

\begin{equation}
\ell(a,b)
=
\left|
\operatorname{wrap}(a-b)
\right|.
\label{eq:circular_error}
\end{equation}

The local visual-increment loss is

\begin{equation}
\mathcal L_{\Delta\psi}
=
\frac{1}{N}
\sum_t
\ell(\widehat{\Delta\psi}_t,\Delta\psi_t),
\label{eq:local}
\end{equation}

where \(N\) is the number of sampled transitions.

To supervise accumulated turn, we use horizons
\(\mathcal H=\{2,4,8\}\). The ground-truth change over \(k\) transitions is

\begin{equation}
\Delta\psi_{t,k}
=
\operatorname{wrap}
\left(
\psi_{t+k}-\psi_t
\right),
\label{eq:gt_path}
\end{equation}

and the path-integration loss is

\begin{equation}
\mathcal L_{\mathrm{PI}}
=
\frac{1}{|\mathcal H|}
\sum_{k\in\mathcal H}
\frac{1}{|\mathcal W_k|}
\sum_{t\in\mathcal W_k}
\ell
\left(
\sum_{\tau=t}^{t+k-1}\widehat{\Delta\psi}_\tau,
\Delta\psi_{t,k}
\right),
\label{eq:path}
\end{equation}

where \(\mathcal W_k\) contains the valid starting indices for horizon \(k\).

The ordered predictions are additionally supervised by

\begin{equation}
\mathcal L_{\mathrm{br}}
=
\frac{1}{N}
\sum_t
\left[
\ell(q_{f,t},\Delta\psi_t)
+
\ell(-q_{r,t},\Delta\psi_t)
\right],
\label{eq:branch}
\end{equation}

and their symmetric component is penalized by

\begin{equation}
\mathcal L_{\mathrm{sym}}
=
\frac{1}{N}
\sum_t
\left|
\frac{q_{f,t}+q_{r,t}}{2}
\right|.
\label{eq:sym}
\end{equation}

Antisymmetry of \(\widehat{\Delta\psi}_t\) follows from
Eq.~\eqref{eq:skew} by construction. The auxiliary losses
\(\mathcal L_{\mathrm{br}}\) and \(\mathcal L_{\mathrm{sym}}\) instead
regularize the ordered predictions before their antisymmetric projection.

The complete objective is
\vspace{-2pt}
\begin{equation}
\mathcal L =
4\mathcal L_{\Delta\psi}
+0.1\mathcal L_{\mathrm{PI}}
+\mathcal L_{\mathrm{br}}
+0.5\mathcal L_{\mathrm{sym}}.
\label{eq:loss}
\end{equation}

No loss is applied directly to \(s_t\); its evolution is fixed by
Eq.~\eqref{eq:update}.

At inference, each new image is paired with the preceding image.
Equations~\eqref{eq:branches}--\eqref{eq:skew} produce the visual increment,
Eq.~\eqref{eq:update} advances the latent state, and
Eq.~\eqref{eq:heading_readout} recovers the heading. RingNav is causal and
requires no future frames, map, keyframes, or learned latent-state transition.
\section{Experiments}
\label{sec:results}

We evaluate RingNav on synthetic benchmarks and real Crazyflie flights, and
organize the experiments around the following questions:

\textbf{Q1.} How accurately does RingNav estimate heading across CompTest,
HDH-Bench, and real data?

\textbf{Q2.} What is the effect of the \(S^1\) heading state, fixed rotation
update, and antisymmetric visual turn?

\textbf{Q3.} How does RingNav behave when visual input is degraded or missing?

\textbf{Q4.} Does RingNav transfer to the real world without fine-tuning?

\textbf{Q5.} What computation and memory does RingNav require on
resource-constrained hardware?

\subsection{Experimental Setup}
\label{sec:setup}

\textbf{Data.}
Training uses 104 sequences: 80 TartanAir trajectories from 13 environments
and 24 rendered trajectories spanning an apartment, a workshop, an urban
alley, and an atrium
\cite{wang2020tartanair,denninger2023blenderproc}. Eight additional TartanAir
trajectories are used for validation. Evaluation uses the 16-trajectory
TartanAir Visual SLAM Challenge test set (CompTest), HDH-Bench with 196
trajectories from four rendered scenes not used for training, and 35
Crazyflie sequences totaling 18,974 image transitions. The four HDH-Bench
scenes comprise a repetitive corridor, a large urban environment, a parking
garage, and a forest. The Crazyflie set contains 23 clean flights and 12
degraded sequences derived from four turning flights.

No real-world data is used for training or model selection. Training does not
contain any blur, flicker, rolling-shutter, lens-contamination, low-light,
low-contrast, camera-freeze, or repeated-frame augmentations.

\textbf{Input processing.}
RingNav uses pairs of \(320\times320\) RGB images. Synthetic \(4{:}3\)
images are resized directly to this resolution. For Crazyflie, a
\(406\times306\) region centered on the calibrated principal point is cropped
and resized to \(320\times320\).

\textbf{Metrics.}
For predicted and ground-truth turns
\(\widehat{\Delta\psi}_t\) and \(\Delta\psi_t\), the turn residual is
\vspace{-2pt}
\begin{equation}
r_t =
\operatorname{wrap}
\left(
\widehat{\Delta\psi}_t-\Delta\psi_t
\right).
\end{equation}

We report mean absolute turn error,
\begin{equation}
E_{\mathrm{MAE}}
=
\frac{1}{N}\sum_t |r_t|.
\end{equation}

To measure accumulated drift, we also report
\begin{equation}
D_{4\mathrm{s}}
=
\frac{1}{M}
\sum_i
\left|
\sum_{t\in W_i} r_t
\right|,
\label{eq:d4s}
\end{equation}

the mean absolute accumulated turn error over four-second sliding windows.
At the nominal sampling rates of \(10\) Hz for CompTest, \(24\) Hz for
HDH-Bench, and approximately \(31.2\) Hz for Crazyflie, four seconds
corresponds to \(40\), \(96\), and \(125\) transitions, respectively.
Sequence metrics are weighted by the number of transitions.

TartanVO, DPVO, and ORB-SLAM2 are each evaluated over five runs.
DROID-SLAM is evaluated over 11 runs on CompTest and five runs on HDH-Bench
and Crazyflie. For each repeated-run baseline, we report the median turn MAE,
median \(D_{4\mathrm{s}}\), and median coverage across runs. If a method does
not return an estimate for a transition, that transition is scored as zero
turn. Methods with median coverage below \(80\%\) are marked \(\times\).
Crazyflie transitions without valid motion-capture measurements are excluded
from turn MAE. For two clean flights shorter than four seconds, accumulated
turn error is computed over the available sequence duration.

For turn scoring, TartanVO uses the \(z\)-component of the rotation vector of
\(R_t^\top R_{t+1}\), while DPVO, DROID-SLAM, and ORB-SLAM2 use the
\(y\)-component in their camera convention. Visual Gyro uses the same
camera-\(y\) turn. Ground-truth heading is world yaw about gravity for
TartanAir, optical-axis azimuth for HDH-Bench, and airframe azimuth for
Crazyflie. No initial-orientation or trajectory alignment is applied before
scoring.

\textbf{Training.}
RingNav and the four model variants mentioned in
Table~\ref{tab:memory}--A are trained for 50 epochs with AdamW, a learning
rate of \(3\times10^{-4}\), weight decay \(3\times10^{-4}\), and batch size
8. Checkpoints are selected by the mean geodesic heading error over the 24
states of each stride-1 validation window, initialized from its ground-truth
starting heading and averaged over the eight validation trajectories.

\subsection{Heading Estimation}
\label{sec:domains}

We compare RingNav with TartanVO, DPVO, DROID-SLAM, and ORB-SLAM2 as
full visual-motion estimators
\cite{wang2021tartanvo,teed2023deep,teed2021droid,
murartal2017orbslam2}, and with the Visual Gyro of
\cite{hartmann2015visualgyro} as a rotation-only estimator. We also include
an always-zero turn baseline. Table~\ref{tab:main} summarizes heading
estimation across CompTest, HDH-Bench, and Crazyflie.

Each baseline retains its own visual front end. On HDH-Bench, TartanVO uses
\(640\times448\) input, DPVO \(640\times480\), DROID-SLAM \(512\times384\),
and Visual Gyro \(320\times320\) grayscale images. On Crazyflie, the full
visual-motion methods use the calibrated full-resolution stream with their
method-specific preprocessing, while Visual Gyro resizes the full frame to
\(320\times320\) grayscale.

\begin{table}[t]
\caption{\textbf{Heading estimation across domains.}
Turn MAE / four-second accumulated turn error \(D_{4\mathrm{s}}\), in
degrees. \(\times\) denotes coverage below \(80\%\); the best result within
each method group is bolded.}
\label{tab:main}
\centering
\footnotesize
\renewcommand{\arraystretch}{1.10}
\setlength{\tabcolsep}{3.1pt}

\begin{tabular*}{\columnwidth}{@{\extracolsep{\fill}}lccc@{}}
\toprule
\textbf{Method}
& \textbf{CompTest}
& \textbf{HDH-Bench}
& \textbf{Crazyflie}\\
\midrule

\multicolumn{4}{c}{\textit{Reference}}\\
\midrule

Zero turn
& 0.861 / 22.3
& 1.741 / 158.7
& 0.774 / 65.2\\

\midrule
\multicolumn{4}{c}{\textit{Full Visual Motion}}\\
\midrule

TartanVO
& 0.206 / 4.8
& 0.267 / 22.9
& 0.347 / 12.1\\

DPVO
& \textbf{0.177 / 4.2}
& \textbf{0.106 / 5.3}
& \textbf{0.336 / 4.3}\\

DROID-SLAM
& 0.198 / 4.6
& 0.148 / 7.0
& 0.343 / 4.4\\

ORB-SLAM2
& \(\times\)
& \(\times\)
& \(\times\)\\

\midrule
\multicolumn{4}{c}{\textit{Rotation / Heading Only}}\\
\midrule

Visual Gyro
& \textbf{0.323 / 6.8}
& \(\times\)
& 0.511 / 26.9\\

\textbf{RingNav}
& 0.589 / 17.6
& \textbf{0.417 / 34.5}
& \textbf{0.458 / 22.2}\\
\bottomrule
\end{tabular*}
\end{table}

\begin{figure}[t]
\centering
\includegraphics[width=0.95\columnwidth]{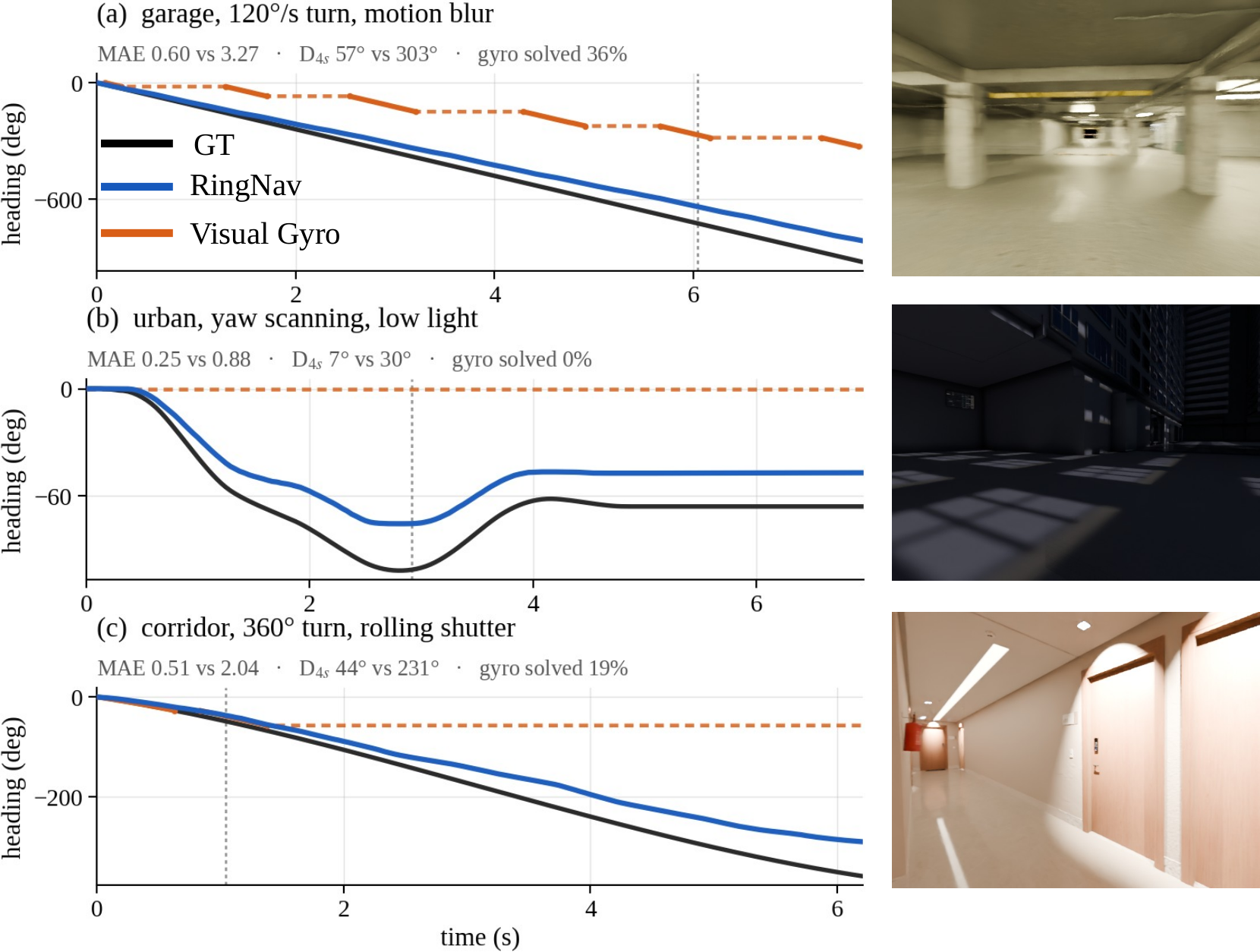}
\caption{\textbf{Heading estimates on HDH-Bench.}
Ground truth, RingNav, and Visual Gyro on representative trajectories.}
\label{fig:yaw196}
\end{figure}

On CompTest, Visual Gyro provides the more accurate rotation estimate,
reaching \(0.323^\circ\) turn MAE and \(6.8^\circ\) four-second accumulated
turn error compared with \(0.589^\circ\) and \(17.6^\circ\) for RingNav.
The full visual-motion estimators are also more accurate on this benchmark;
RingNav instead targets a compact heading-only estimator whose computational
cost is evaluated in Sec.~\ref{sec:deploy}.

Across the 196 HDH-Bench trajectories, RingNav completes the full set with
\(0.417^\circ\) turn MAE and \(34.5^\circ\) four-second accumulated turn
error. The scenes contain repetitive indoor structure, large outdoor
environments, a parking garage, and a forest, none of which appears in
training. Representative heading estimates are shown in
Fig.~\ref{fig:yaw196}. Visual Gyro falls below the \(80\%\) coverage threshold
on this set.

Across the 35 Crazyflie sequences, RingNav reaches \(0.458^\circ\) turn MAE
and \(22.2^\circ\) four-second accumulated turn error, compared with
\(0.511^\circ\) and \(26.9^\circ\) for Visual Gyro at \(83.4\%\) pooled
coverage. RingNav also improves over the zero-turn baseline on all three
evaluation sets. Section~\ref{sec:real} separates clean and degraded flights.

\subsection{State and Update Ablation}
\label{sec:memlaw}

We next isolate the contribution of the persistent state, its update, and the
visual turn using four similarly sized variants, summarized in
Table~\ref{tab:memory}. PairGRU learns a recurrent state and its update.
learned-\(S^1\) keeps the heading state on \(S^1\) but learns the update.
exact-\(S^1\) uses the \(S^1\) state and fixed rotation update from
Eq.~\eqref{eq:update}, but predicts the visual turn from one image order.
skew-\(S^1\) adds the antisymmetric visual turn from Eq.~\eqref{eq:skew},
but omits the branch and symmetry losses. RingNav uses the complete model in
Sec.~\ref{sec:method}.

RingNav, exact-\(S^1\), and skew-\(S^1\) contain 10,600 trainable
parameters; learned-\(S^1\) contains 9,642 and PairGRU 10,890. Each model
also contains four frozen parameters in a \(2\times2\) identity layer used
for state initialization. All five models use the same training data and
lightweight visual encoder. The four \(S^1\) variants carry an 8-byte
persistent state, compared with the 64-byte, 16-dimensional state of
PairGRU.

For the forward--reverse test, each of the 16 CompTest sequences is traversed
forward and backward repeatedly for up to \(10^5\) transitions.
\(E_{\rm rev}\) is the transition-weighted mean absolute unwrapped accumulated
turn. \(E_{\rm mem}(N)\) measures state error after \(N\) updates when the
same world-heading turn sequence is applied to each update model.

\begin{table}[t]
\caption{\textbf{Ablation of the latent state, update, and visual increment.}
(A) End-to-end heading estimation and forward--reverse error.
(B) State error after repeated updates with the same turn sequence.}
\label{tab:memory}
\centering
\scriptsize
\renewcommand{\arraystretch}{1.10}
\setlength{\tabcolsep}{2.6pt}

\begin{tabular*}{\columnwidth}
{@{\extracolsep{\fill}}lccrrr@{}}
\toprule
\multicolumn{6}{c}{\textbf{A. End-to-End Heading Estimation}}\\
\midrule
\textbf{Variant}
& \textbf{State}
& \textbf{Update}
& \textbf{Turn MAE}
& \textbf{\(D_{4\mathrm{s}}\)}
& \textbf{\(E_{\rm rev}\) (\(^{\circ}\))}\\
\midrule

PairGRU
& learned & learned
& 0.810 & 20.6 & 53.4\\

learned-\(S^1\)
& \(S^1\) & learned
& 0.636 & 19.6 & 3,205\\

exact-\(S^1\)
& \(S^1\) & fixed
& 0.641 & 19.7 & 6,190\\

skew-\(S^1\)
& \(S^1\) & fixed
& 0.607 & 17.9 & 0\\

\textbf{RingNav}
& \(S^1\) & fixed
& \textbf{0.589}
& \textbf{17.6}
& \textbf{0}\\
\bottomrule
\end{tabular*}

\vspace{6pt}

\begin{tabular*}{\columnwidth}
{@{\extracolsep{\fill}}lrrr@{}}
\toprule
\multicolumn{4}{c}{
\textbf{B. State-Update Test: \(E_{\rm mem}(N)\) (\(^{\circ}\)) \(\downarrow\)}
}\\
\midrule
\textbf{Update}
& \(\mathbf{10^3}\)
& \(\mathbf{10^4}\)
& \(\mathbf{10^5}\)\\
\midrule

learned MLP, \(\mathbb R^2\)
& 72.2 & 70.5 & 71.8\\

learned GRU, \(\mathbb R^{16}\)
& 36.1 & 37.3 & 55.7\\

\textbf{fixed rotation}, FP32
& \(3.1{\times}10^{-5}\)
& \(1.5{\times}10^{-4}\)
& \(\mathbf{4.6{\times}10^{-4}}\)\\
\bottomrule
\end{tabular*}
\end{table}

\begin{figure*}[t!]
\centering
\includegraphics[width=0.92\textwidth]{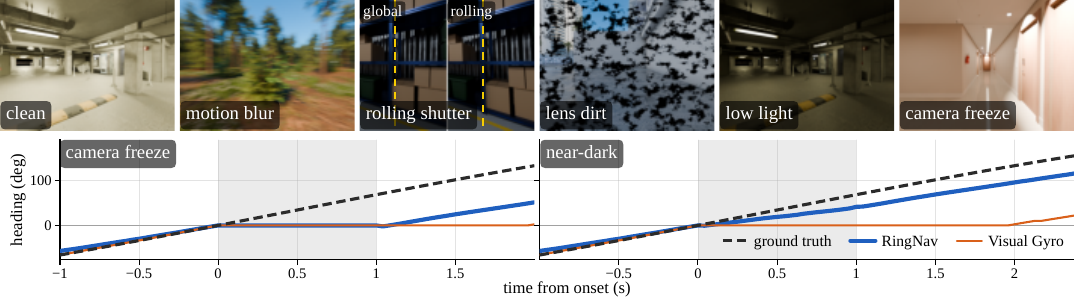}
\caption{\textbf{Heading under degraded and missing visual input.}
Example inputs and heading estimates under image degradation, camera freeze,
and near-dark observations.}
\label{fig:degradation}
\end{figure*}

\begin{table*}[t!]
\caption{\textbf{Heading under degraded and missing visual input.}
(A) Turn MAE / four-second accumulated turn error under image degradation.
(B) Predicted turn magnitude, accumulated turn error over the test interval,
and local residual RMSE over the following 20 transitions.}
\label{tab:robust}
\centering
\footnotesize
\renewcommand{\arraystretch}{1.06}

\begin{minipage}[t]{0.485\textwidth}
\centering
\textbf{A. Degraded Visual Input}\\[2pt]

\scriptsize
\setlength{\tabcolsep}{2.3pt}
\begin{tabular*}{\linewidth}{@{\extracolsep{\fill}}lrrrrr@{}}
\toprule
\textbf{Method}
& \textbf{Clean}
& \textbf{Blur}
& \textbf{Flicker}
& \textbf{Roll.\ shutter}
& \textbf{Lens dirt}\\
\midrule

\multicolumn{6}{c}{\textit{Full Visual Motion}}\\
\midrule

TartanVO
& 0.105/7.9 & 0.104/8.0 & 0.105/7.9 & 0.107/7.9 & 0.529/16.4\\

DPVO
& 0.021/0.12 & 0.022/0.13 & 0.021/0.12 & 0.022/0.17 & \(\times\)\\

DROID-SLAM
& 0.069/1.00 & 0.076/1.0 & 0.023/0.18 & 0.049/0.99 & 0.625/16.0\\

\midrule
\multicolumn{6}{c}{\textit{Rotation / Heading Only}}\\
\midrule

Visual Gyro
& 0.141/5.8 & \(\times\) & 0.151/6.0 & 0.142/5.9 & 0.535/15.8\\

\textbf{RingNav}
& 0.143/6.9 & 0.137/6.9 & 0.291/8.9 & 0.142/6.9 & 0.408/16.1\\
\bottomrule
\end{tabular*}
\end{minipage}
\hfill
\begin{minipage}[t]{0.485\textwidth}
\centering
\textbf{B. Missing and Weak Visual Input}\\[2pt]

\scriptsize
\setlength{\tabcolsep}{1.9pt}
\begin{tabular*}{\linewidth}{@{\extracolsep{\fill}}lrrrrrr@{}}
\toprule
& \multicolumn{3}{c}{\textbf{Camera Freeze}}
& \multicolumn{3}{c}{\textbf{Near-Dark}}\\
\cmidrule(lr){2-4}\cmidrule(lr){5-7}

\textbf{Method}
& \(\overline{|\widehat{\Delta\psi}|}\)
& \textbf{Win.}
& \textbf{Post-20}
& \(\overline{|\widehat{\Delta\psi}|}\)
& \textbf{Win.}
& \textbf{Post-20}\\
\midrule

\multicolumn{7}{c}{\textit{Full Visual Motion}}\\
\midrule

TartanVO
& 0.008 & 40.75 & 0.362
& 1.538 & 4.62 & 0.362\\

DPVO
& 0.013 & 40.77 & 0.534
& 1.701 & 0.47 & 0.068\\

DROID-SLAM
& 0.275 & 42.79 & 0.226
& 1.745 & 1.81 & 0.306\\

\midrule
\multicolumn{7}{c}{\textit{Rotation / Heading Only}}\\
\midrule

Visual Gyro
& \(\times\) & \(\times\) & \(\times\)
& \(\times\) & \(\times\) & \(\times\)\\

\textbf{RingNav}
& 0.000 & 40.76 & 0.355
& 1.114 & 16.46 & 0.355\\
\bottomrule
\end{tabular*}
\end{minipage}
\end{table*}

Restricting the persistent state to \(S^1\) reduces turn MAE from
\(0.810^\circ\) for PairGRU to \(0.636^\circ\), while four-second accumulated
turn error decreases from \(20.6^\circ\) to \(19.6^\circ\). Replacing the
learned update with fixed rotation changes local accuracy little:
exact-\(S^1\) reaches \(0.641^\circ\) turn MAE and \(19.7^\circ\)
accumulated turn error. Thus, fixing the known state update removes a learned
degree of freedom, but does not by itself improve visual turn estimation.

The antisymmetric visual turn affects a different property. Variants without
the order-reversal constraint accumulate a nonzero residual under repeated
forward and reverse traversal, while skew-\(S^1\) and RingNav give
\(E_{\rm rev}=0\). RingNav also gives the lowest turn MAE and four-second
accumulated turn error among the five compact models. The exact constraint
requires two ordered evaluations: one \(320\times320\) RGB pass costs
\(20.74\) MMAC, compared with \(41.48\) MMAC per RingNav update.

Panel B isolates the state update from the image network. With the same
world-heading turn parameterization, the learned updates accumulate tens of
degrees of state error over long sequences, while fixed FP32 rotation remains
near zero after \(10^5\) updates.

\vspace{-2pt}
\subsection{Degraded and Missing Visual Input}
\label{sec:robust}

We then test how the estimators behave when visual information is degraded,
repeated, or weak. Table~\ref{tab:robust}--A reports turn MAE and
\(D_{4\mathrm{s}}\), while Fig.~\ref{fig:degradation} shows representative
inputs and heading estimates. The controlled degradations use a
\(0.5\)-frame blur shutter, periodic flicker, a \(0.3\)-frame top-to-bottom
rolling-shutter readout with a \(0.5\)-frame shutter, and lens contamination
covering \(34.6\%\) of the image. Table~\ref{tab:robust}--B uses 24
byte-identical frames for camera freeze and a \(7.64\)-stop exposure
reduction for near-dark input.

In Table~\ref{tab:robust}--B,
\(\overline{|\widehat{\Delta\psi}|}\) is mean predicted turn magnitude,
Win.\ is accumulated turn error over the test interval, and Post-20 is the
local residual RMSE over the following 20 transitions. All values are in
degrees, except predicted turn magnitude, which is in degrees per transition.

Under blur and rolling shutter, RingNav remains close to its clean result.
Flicker and lens contamination increase its error. Under lens contamination,
RingNav gives the lowest local turn error among methods that return estimates,
but not the lowest four-second accumulated turn error. These results do not
show general robustness to image degradation.

Camera freeze tests the identical-frame property in
Eq.~\eqref{eq:identities}. During camera freeze, consecutive delivered frames
are byte-identical while the platform continues to rotate. RingNav therefore
predicts exactly zero turn and accumulates \(40.76^\circ\) of error over the
freeze interval. Once image change returns, local turn estimation resumes,
but the accumulated heading offset remains.

The complementary stationary case tests whether identical observations create
spurious motion when the true turn is zero. Across four such sequences,
PairGRU, learned-\(S^1\), and exact-\(S^1\) accumulate \(2.04^\circ\),
\(2.28^\circ\), and \(2.69^\circ\) of false turn, while skew-\(S^1\) and
RingNav remain at zero. Together, the stationary and turning cases show both
the benefit and limitation of the constraint: identical observations cannot
introduce a visual turn, but they also cannot reveal motion during a camera
freeze. RingNav does not currently distinguish these cases with a validity
signal.

Near-dark images retain weak visual structure. RingNav therefore produces
nonzero turn estimates and accumulates \(16.46^\circ\) error over the
near-dark interval, while DPVO remains substantially more accurate. Together,
the camera-freeze and near-dark tests separate the exact behavior under
identical observations from estimation when visual information is weak but
nonzero.

\subsection{Real-Flight Transfer}
\label{sec:real}

\begin{figure}[t!]
\centering
\includegraphics[width=\columnwidth]{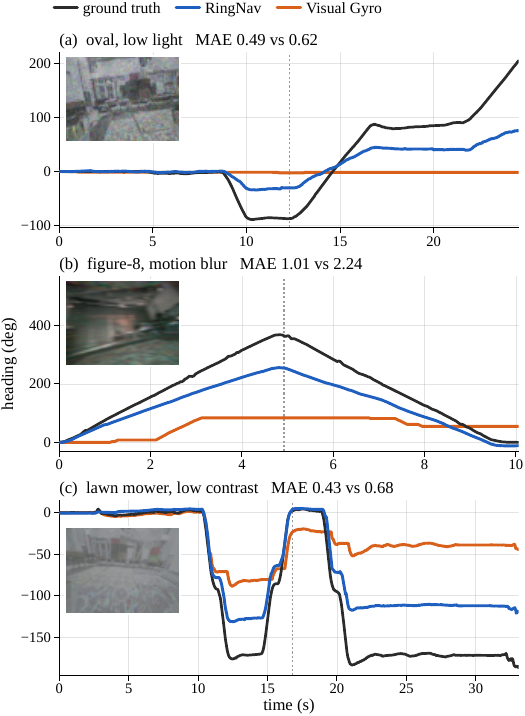}
\caption{\textbf{Heading on real Crazyflie flights.}
Ground truth, RingNav, and Visual Gyro on representative degraded sequences.}
\label{fig:crazyflie_heading}
\end{figure}

We evaluate zero-shot transfer using the frozen RingNav checkpoint on 23
recorded Crazyflie flights and 12 degraded sequences derived from four of
those flights, for a total of 18,974 image transitions at approximately
\(31.2\) Hz. The \(640\times480\) FPV stream is synchronized with
motion-capture heading. No real imagery is used for training or model
selection, and all real-flight inference is performed offline on the recorded
stream. Representative degraded-flight heading estimates are shown in
Fig.~\ref{fig:crazyflie_heading}.

The 12 degraded sequences apply one of three controlled changes to four
turning flights: \(33\)-pixel motion blur, a \(7\)-stop simulated
underexposure followed by gain compensation to nominal brightness, preserving
the resulting low-light noise, or \(0.3\times\) contrast.

On the 23 clean flights, Visual Gyro provides the more accurate estimate,
reaching \(0.325^\circ\) turn MAE and \(3.3^\circ\) four-second accumulated
turn error, compared with \(0.379^\circ\) and \(11.6^\circ\) for RingNav.

Under degradation, Visual Gyro coverage falls to \(79\%\) with motion blur,
\(4\%\) under low light, and \(78\%\) under reduced contrast, while RingNav
returns an estimate for every image pair. Across the 12 degraded sequences,
RingNav reaches \(0.602^\circ\) turn MAE and \(41.9^\circ\) four-second
accumulated turn error. Visual Gyro has \(53.8\%\) coverage and therefore
falls below the reporting threshold. Scoring its missing updates as zero
turns gives \(0.853^\circ\) turn MAE and \(70.3^\circ\) accumulated turn
error. A zero-turn baseline reaches \(1.034^\circ\) and \(100.0^\circ\).

These results show a different operating point rather than uniform accuracy
improvement over Visual Gyro. The feature-based estimator is more accurate on
clean flights and can remain more accurate when sufficient feature coverage
is available, while RingNav continues returning estimates across all three
tested degradations. Section~\ref{sec:deploy} evaluates the computational
cost of the two estimators on edge hardware.

\subsection{Edge Deployment}
\label{sec:deploy}

Finally, we benchmark RingNav on three Raspberry Pi platforms and on the
GAP8 processor of the Crazyflie AI-deck. Visual Gyro is evaluated on the
Raspberry Pi platforms for comparison. Table~\ref{tab:deploy} reports
measured heading-update rate and peak runtime memory. All reported rates
exclude camera capture.

\begin{table}[t!]
\caption{\textbf{Edge deployment across hardware platforms.}
Heading-update rate and peak runtime memory on different embedded boards.}
\label{tab:deploy}
\centering
\scriptsize
\renewcommand{\arraystretch}{1.12}
\setlength{\tabcolsep}{2.6pt}

\begin{tabular*}{\columnwidth}{@{\extracolsep{\fill}}lrrrr@{}}
\toprule
&
\multicolumn{2}{c}{\textbf{RingNav}}
&
\multicolumn{2}{c}{\textbf{Visual Gyro}}\\
\cmidrule(lr){2-3}
\cmidrule(lr){4-5}
\textbf{Platform}
& \textbf{RAM (MiB)}
& \textbf{Hz}
& \textbf{RAM (MiB)}
& \textbf{Hz}\\
\midrule

Pi 5
& 73.8 & 199
& 104.5 & 3.50\\

Pi 4
& 68.4 & 63.3
& 102.1 & 1.21\\

Pi Zero 2W
& 68.2 & 27.3
& 130.2 & 0.349\\

GAP8 (AI-deck)
& 0.49 & 33.4
& 7.83& 0.1\\

\bottomrule
\end{tabular*}
\end{table}

In the measured Raspberry Pi implementations, RingNav provides a
\(52\)--\(78\times\) higher heading-update rate than Visual Gyro. This
complements the accuracy results in Secs.~\ref{sec:domains}
and~\ref{sec:real}: Visual Gyro is more accurate on clean imagery, while
RingNav targets the computational regime of resource-constrained robots.

The GAP8 deployment uses \(320\times240\) grayscale input and is reported
separately from the \(320\times320\) RGB configuration used in the accuracy
experiments. RingNav reaches \(33.4\) Hz and the complete runtime fits in
\(0.49\) MiB of on-chip memory. The persistent heading state itself occupies
8 bytes. The current GAP8 implementation of Visual Gyro requires external
HyperRAM and is therefore not included in the on-chip comparison.

\vspace{-3pt}
\section{Conclusion}

Tiny robots need heading, yet most visual estimators recover substantially
more state than the task requires. RingNav instead fixes the persistent
heading state to \(S^1\), updates it by an \(SO(2)\) rotation, and learns only
the visual turn between consecutive frames. This structured memory improves
accuracy over a similarly sized learned-memory model, while antisymmetry gives
exact sign reversal under image-order reversal and zero predicted turn for
identical images.

RingNav transfers from synthetic training to real Crazyflie flight data
without fine-tuning and continues returning estimates under the tested
motion-blur, low-light, and low-contrast degradations. It runs at
\(33.4\,\mathrm{Hz}\) on GAP8 with the complete runtime contained in
\(0.49\,\mathrm{MiB}\) of on-chip memory.

Its main limitation remains visual turn accuracy: the fixed geometric update
cannot correct an inaccurate or unobservable increment. Future work will
improve the visual estimator while preserving the compact structured state,
and extend task-shaped memory to richer navigation manifolds.


{\footnotesize
\bibliographystyle{ieeetr}
\bibliography{ref}
}

\end{document}